\documentclass{IEEEtran}
\usepackage[utf8]{inputenc}
\usepackage{graphicx}
\usepackage{adjustbox}
\usepackage{color, colortbl}
\definecolor{chartreuse(traditional)}{rgb}{0.87, 1.0, 0.0}
\title{A Novel Pipeline for Improving Optical Character Recognition through Post-processing Using Natural Language Processing}
\author{\IEEEauthorblockN{Aishik Rakshit$^1$, Samyak Mehta$^2$, and Anirban Dasgupta$^3$}\\
\IEEEauthorblockA{\textit{Department of Electronics and Electrical Engineering} \\
\textit{Indian Institute of Technology Guwahati, India}\\
$^1$a.rakshit@iitg.ac.in,$^2$ samyak.sanjay@iitg.ac.in, $^3$anirban.dasgupta@iitg.ac.in}
}
\begin{document}
\maketitle
\begin{abstract}
    Optical Character Recognition (OCR) technology finds applications in digitizing books and unstructured documents, along with applications in other domains such as mobility statistics, law enforcement, traffic, security systems, etc. The state-of-the-art methods work well with the OCR with printed text on license plates, shop names, etc. However, applications such as printed textbooks and handwritten texts have limited accuracy with existing techniques. The reason may be attributed to similar-looking characters and variations in handwritten characters. Since these issues are challenging to address with OCR technologies exclusively, we propose a post-processing approach using Natural Language Processing (NLP) tools. This work presents an end-to-end pipeline that first performs OCR on the handwritten or printed text and then improves its accuracy using NLP.
\end{abstract}
\section{Introduction}
Optical Character Recognition (OCR) is a technology for extracting texts from images containing text information \cite{liang2005camera}. Such images occur from photos containing text information, scanned documents, scene photos, subtitle text superimposed on an image, etc.
OCR is useful as images consume more memory space than text files. Moreover, text information is easier to copy and edit and helpful in many artificial intelligence (AI) tools, particularly for Natural Language Processing (NLP) problems.
Some general applications include self-service utility meter reading, intelligent traffic surveillance and parking system, license plate recognition, contactless check-in at private and public transportation stations, intelligent security systems, digitizing old books, etc. \cite{bansal2020building}. As such, OCR helps to reduce crime, increase police efficiency, and improve safety \cite{bansal2020building}. 
The OCR methods recognize characters in the image independently by image segmentation considering only the shape and structure of the characters.
Significant research on OCR has been reported on recognizing texts from scanned documents, and number plates, with sufficient performance. Even OCR on handwritten texts in different languages has received much attention, however, with limited accuracy. 
Hence, there is scope for improvement in the efficiency of OCR of handwritten text. Even the OCR of printed text is yet to be perfect. The prime challenges for inaccurate or missing text are as follows:
\begin{itemize}
    \item variations in font style and size,
    \item case sensitivity,
    \item similar character shapes, such as `o' and `0',
    \item varying orientations.
\end{itemize}
These OCR mistakes negatively impact several NLP applications, including text summarizing, part-of-speech (POS) tagging, sentence boundary detection, topic modeling, named entity recognition (NER), and text classification.
The ability of NER tools to detect and identify proper nouns and classify them into the person, place, and organization categories significantly deteriorates when the error rate (ER) of OCR output rises. Post-processing OCR outputs can significantly help correct these mistakes and increase the accuracy of the outputs.\\
Hence, the objective is to develop an end-to-end pipeline that first performs OCR on the single-line handwritten or printed text and then improves its accuracy by post-processing the OCR output using NLP.
\subsection{Prior Art}
The current OCR approaches use Convolutional Neural Network (CNN)-based encoders for picture interpretation and Recurrent Neural Network (RNN)-based decoders for text generation.
The two most popular OCR models are the Transformer-based OCR (Tr-OCR) model \cite{DBLP:journals/corr/abs-2109-10282} and the Paddle-Paddle OCR (PP-OCR) model \cite{DBLP:journals/corr/abs-2009-09941}.
The Tr-OCR model uses the Transformer architecture for workpiece-level text generation and image understanding. TrOCR has a pre-trained image Transformer as an encoder with the decoder as a pre-trained text Transformer. This model has been trained on the IAM handwritten dataset. \\
The PP-OCR model consists of text detection, text recognition, and detected box rectification using a convolutional recurrent neural network (CRNN) as a text recognizer at the back end. The CRNN has convolutional layers for feature extraction followed by recurrence for sequence modeling.
These architectures produce efficient results if trained on a specific type of data. However, generalizing is difficult on unconstrained datasets due to the large variability.\\
In the domain of OCR output correction, the prior algorithms used mainly operate on the standard pipeline with the delete operation, followed by transposes, followed by replaces, and finally, inserts. This method used in implementing text blob's spelling correction has taken Peter Norvig's "How to Write a Spelling Corrector" \cite{norvig2007write} as ground truth for training. This approach is improved using Symspellpy \cite{lansley2020seader++}. The symmetric delete spelling correction algorithm lowers the complexity of edit candidate generation and dictionary lookup for a specific Damerau-Levenshtein distance. It is language-independent and is about six times faster than the traditional approach.
\section{Materials and Methods}
This firstly evaluates two OCR models in this work, \textit{viz.}, Tr-OCR and PP-OCR, on various handwritten and printed datasets.
Thsi work then choose the better-fitting model for recognizing single-line handwritten text. A line segmentation module for segmenting a multi-line document into single lines and a classifier that classifies each of these single lines into printed or handwritten text are also implemented. The output of the OCR model is then fed to our post-processing model, which improves the accuracy of the OCR output. \\
The OCR output post-processing task aims to identify the sequence of words $X = x_1 x_2... x_m$ present in the original hardcopy document given a sequence of n OCR degraded tokens $Y = y_1 y_2... y_n$. It should be noted that $n$ and $m$ are not always equal because segmentation errors could result in OCR sub-sequences that are not correct word sequences.\\
We divide our work into two modules. The first consists of the segmentation unit, the classification unit and the OCR model unit. The OCR models are evaluated on various real-life datasets. We then select the better-fit model as input to the second module, i.e., NLP-based post-processing. This module takes in the outputs of the OCR model and then post-processes it using NLP techniques to minimize error.
\subsection{Module-A: OCR Engine}
Module A consists of the first half of the pipeline which is to first perform line segmentation on a multi-line document, then classify each line into printed and handwritten text using a classifier, then perform OCR on it using a suitable OCR model.
Evaluation has been performed on two existing popular OCR models on various datasets with different fonts, handwritten dataset, dataset with occluded or background color and noise.

\subsubsection{Segmentation}
The aim here is to Segmenting lines in documents using A* Path planning algorithm \cite{segmentation}.
The method to achieve this is to:
\begin{enumerate}
    \item We first input a non-skewed document of either handwritten or printed text into this model and then convert the input image to 2D grayscale image.
    \item We use sobel filter to detect the text edges in the image. The image is convolved with two 3*3 kernels (horizontal and vertical), to calculate the image derivatives.
    \item We then find the horizontal projection profile of the edge detected image. HPP is calculated by the array of the sum of elements in each row. So more peaks will be seen corresponding to the rows that have text whereas the blank areas will not peak in the HPP graph.
    \item We then detect peaks, for which I take the threshold of one-fourth difference of maximum hpp and minimum hpp value. This helps in dividing the potential line segment regions from the text.
    \item We then make a cut in places where upper line text connects with the lower line text.
    \item We then use the A* path planning along the segmentation region and record the paths. This helps in segmenting the document into single lines.
    \end{enumerate}

\subsubsection {classification}
Convolutional neural networks (CNN) are used to classify text lines as either printed or handwritten, however it is actually the collection and preparation of the data that presents the biggest challenges. Presenting enough samples to an artificial neural network (ANN) is sufficient to achieve a decent level of accuracy for a wide range of tasks. In fact, current artificial neural networks (ANN) are already capable of handling extremely complicated data (such as ImageNet, which includes 90 different dog breeds to discriminate).
\\The system we created for this work is a DenseNet-121 that has been modified for the binary classification of handwritten and printed text. It is wrapped in some utility classes. A convolutional neural network called DenseNet-121 has 121 layers, the majority of which are tightly connected in 4 blocks. However, compared to designs with more parameters, it has a comparatively low number of parameters for a network of its size and so requires less training data. More information on the classifier used can be found in \cite{classification}.
\subsubsection{Datasets}
The specific datasets that we have used for the purpose are:
\begin{itemize}
    \item Born-Digital Images Dataset \cite{digital}: This dataset contains images made digitally employing a desktop scanner, a camera, and screen capture software. It has 3564 images of words clipped from the actual images and a text file containing the ground truth transcription of all images provided.
    \item Incidental Scene Text Dataset \cite{ch2017total}: This dataset consists of 4468 cut-out word images corresponding to the axis-oriented bounding boxes of the words provided and a single text file with the ground truth.
    \item License Plate Dataset \cite{LP}: This dataset has 209 cropped license plates using the original bounding boxes and has all the single characters labeled, creating a total of 2026 character bounding boxes. Every image comes with a .xml annotation file.
    \item Single Line Handwritten Text Dataset \cite{single}: This dataset \cite{single} contains images of handwritten single-line English texts whose labels are similar to the IAM dataset. There are around 400 images along with their labels.
    \item Bing Images of Short Quotes: This dataset contains about 215 images of short quotes with different background styles. This dataset is unlabelled as its primary purpose is to see the improvements in the outputs after post-processing using NLP.
\end{itemize}
\subsubsection{Performance Metrics}
 The performance evaluations used are character error rate (CER) and word error rate (WER) evaluation metrics. The CER gives the fraction of the number of characters correctly identified, including spaces. The WER is the fraction of the number of words correctly output in reference to the ground truth text.
 \subsection{Module-B: NLP Engine}
 The models we consider are as follows:
 \subsubsection{ByT5}
 The Google AI team debuted T5 \cite{raffel2020exploring}, also known as a Text-To-Text Transfer Transformer, in 2020. The encoder-decoder structure of the T5 transformer model is identical to that of conventional transformer models. There are 12 pairs of blocks of encoder-decoders in it. Self-attention, a feed-forward network, and optional encoder-decoder attention are all present in each block.
 The ByT5 \cite{xue2022byt5} proposes a new model that can directly process raw text, i.e., it would be token-free. The benefits are as follows:
    \begin{itemize}
        \item They can process text in any language. Tokenizers tailored to specific languages are not necessary.
        \item They reduce the trouble of having complicated text preparation pipelines and are noise-resistant.
        \item Now that we only need 256 embeddings for a byte-level model, we no longer need a large vocabulary matrix.
    \end{itemize}
 \subsubsection{BART}
Bidirectional and Auto-Regressive Transformer \cite{lewis2019bart}\\
BART is a pretraining denoising autoencoder for sequence-to-sequence models. The text is first corrupted using a random noise function, and then a model is learned to recreate the original text to train the BART model. It employs a typical Tranformer-based neural machine translation architecture that, despite its simplicity, generalizes several more modern pretraining approaches, including GPT with its left-to-right decoder and BERT (owing to the bidirectional encoder).\\ 
The dataset used to train the models in a supervised manner was generated synthetically from the OSCAR Corpus.
\subsubsection{Alpaca-LORA}
The Alpaca model was optimized through fine-tuning from Meta's LLaMA 7B model, which was achieved through supervised learning on a set of 52K instruction-following demonstrations generated from OpenAI's text-davinci-003. The process of generating the dataset resulted in 52K distinct instructions and corresponding outputs, and was accomplished at a cost of less than \$500 by utilizing the OpenAI API. Hugging Face's training framework was used to fine-tune the LLaMA models, with techniques such as Fully Sharded Data Parallel and mixed precision training being employed. The fine-tuning process of a 7B LLaMA model was accomplished in 3 hours, using 8 80GB A100s. 
We used the Alpaca model in a zero shot manner and it was run in 8-bit precision using bits and bytes.
We tried multiple prompts with the Alpaca-LORA 7B model and the one that worked the best for us was $\textrm{f"Fix all the errors in the sentence : {text}"}$
\subsubsection{Synthetic Dataset Generation}
OCR degraded text is generated for training our byT5 Transformer model using the \textbf{nlpaug} \cite{ma2019nlpaug} library. The OCR Augmentor is used, which can be used to generate character-level errors in the text of the OSCAR \cite{2022arXiv220106642A} Corpus.
\begin{table}[h!]
\centering
\caption{Dataset Generation Parameters Arguments}
\label{table: Generation Parameters}
\begin{tabular}{||c c||} 
 \hline
 Arguement & Value\\ [0.5ex] 
 \hline\hline
 Character Augmentation Probability & 0.7 \\
 Word Augmentation Probability & 0.8\\[1ex] 
 \hline
\end{tabular}
\end{table}

\begin{figure}[htp]
    \centering
    \includegraphics[width=\linewidth]{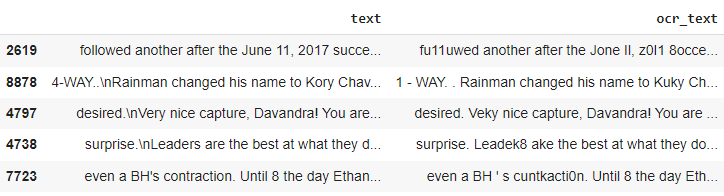}
        \caption{Synthetic Dataset Example}
    \label{fig:example2}
\end{figure}

\subsubsection{Preprocessing Inputs}
To prevent any discrepancies in the lengths of the original Text and the Text Generated by the model with the ground truth, we chunk the texts into lengths of 128 words, but as subword tokenization is being used, we set the max length to 256 but replace all the padding tokens with -100 to prevent loss calculation for them.

\subsubsection{Post-Processing Model Outputs}
The correct spacing insertion into the output from the model is performed using the output distribution. Given a text corpus, we assume that all words are dispersed separately. The relative frequency of each term is then all that is required to know. It is logical to take that they adhere to Zipf's law \cite{piantadosi2014zipf}, which states that the probability of a word having rank $n$ in a list of words is approximate $\frac{1}{n\log N}$, where $N$ refers to the total number of words in the corpus.

After the model is fixed, we can utilize dynamic programming to determine the spaces' locations. The sentence that maximizes the product of the probabilities of each individual word is the most likely one, and dynamic programming makes it simple to calculate. We use a cost defined as the logarithm of the probability's inverse to prevent overflows rather than utilizing the probability itself.
This has been done using the word ninja \cite{wordninja} library.
\section{Results}
\subsection{OCR model evaluation}
We will first discuss the results of the two OCR systems (PP-OCR and Tr-OCR) on various datasets, as discussed above, without any postprocessing. We then proceed to show results of the segmentation and classification sub-modules.

\subsubsection{Dataset 1: Born-Digital Images Dataset}
\begin{figure}
    \begin{center}
	\includegraphics[scale = 1]{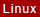}
    \hspace{5pt}
    \includegraphics[scale = 1]{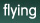}
    \hspace{5pt}
    \includegraphics[scale = 1]{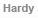}
     \end{center}
	\caption{Sample Born-Digital Images}
	\label{fig: Born-Digital Image}
\end{figure}

The outputs of some sample images in Fig. \ref{fig: Born-Digital Image} are shown in Table \ref{tab: Results on some Born-Digital Images}.
\\
The ultra-weight PP-OCR model, pre-trained in English and Chinese languages, resulted in a CER of 0.44.
While the Tr-OCR model was fine-tuned on the SROIE printed text dataset, it resulted in a CER of 0.3.
Hence Tr-OCR performed better than PP-OCR on this dataset.

\begin{table}[]
    \centering
    \caption{Results on some Born-Digital Images}
    \label{tab: Results on some Born-Digital Images}
    \begin{tabular}{|c|c|c|c|}
    \hline
     Image & Ground Truth & PP-OCR output & Tr-OCR output\\ [0.5ex] 
     \hline\hline
      Sample image 1 & Linux & Linuv & Linux \\
      \hline
       Sample image 2 & flying & flving & tiving \\
      \hline
      Sample image 3 & Hardy & Hordh & Hardy \\ 
      \hline
    \end{tabular}
\end{table}

\subsubsection{Dataset 2: Incidental Scene Text Dataset}
\begin{figure}
    \begin{center}
	\includegraphics[scale = 0.75]{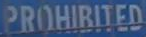}
    \hspace{5pt}
    \includegraphics[scale = 0.75]{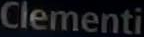}
    \hspace{5pt}
    \includegraphics[scale = 1]{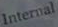}
     \end{center}
	\caption{Sample Incidental Scene Text Images}
	\label{fig: Sample Incidental Scene Text Images}
\end{figure}

The outputs of some sample images in Fig. \ref{fig: Sample Incidental Scene Text Images} are shown in Table \ref{tab: Results on some Incidental Scene Text Images}.
Using the ultra-weight PP-OCR model, which is pre-trained in English and Chinese languages, resulted in a CER of 0.65, while the Tr-OCR model fine-tuned on the SROIE dataset (printed text) resulted in a CER of 0.41.
Hence Tr-OCR performed better than PP-OCR on this dataset.
\\

\begin{table}[]
    \centering
    \caption{Results on some Incidental Scene Text Images}
    \label{tab: Results on some Incidental Scene Text Images}
    \begin{tabular}{|c|c|c|c|}
    \hline
     Image & Ground Truth & PP-OCR output & Tr-OCR output\\ [0.5ex] 
     \hline\hline
      Sample image 1 & PROHIBITED & PROHIRITED & PROHRITED \\[1ex]
      \hline
       Sample image 2 & Clementi & ment & Cementi\\
      \hline
      Sample image 3 & Internal & Interna & Internal \\ 
      \hline
    \end{tabular}
    \vspace{5pt}
\end{table}

\begin{figure}
    \begin{center}
	\includegraphics[scale = 0.8]{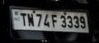}
    \hspace{5pt}
    \includegraphics[scale = 0.5]{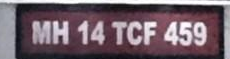}
    \vspace{5pt}
    \includegraphics[scale = 0.5]{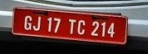}
     \end{center}
	\caption{Sample License Plate Images}
	\label{fig: Sample License Plate Images}
\end{figure}

\begin{table}[h]
    \centering
        \caption{Results on some License Plate Images}
    \label{tab: Results on some License Plate Images}
    \begin{tabular}{|c|c|c|c|}
    \hline
     Image & Ground Truth & PP-OCR output & Tr-OCR output\\ [0.5ex] 
     \hline\hline
      Sample image 1 & TN74F3339 & TN74F339 & TN74F3339 \\[1ex]
      \hline
       Sample image 2 & MH14TCF459 & MH14TCF459 & MN14TCF459\\
      \hline
      Sample image 3 & GJ17TC214 & GJ17TC214 & GJ17TQ214 \\ 
      \hline
    \end{tabular}
\end{table}

\begin{figure}
    \begin{center}
	\includegraphics[width = \linewidth]{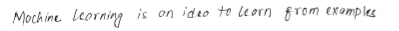}
     \end{center}
     \caption{Sample Single Line Handwritten text Image 1}
     \label{fig: Sample Single Line Handwritten text Image 1}
     \vspace{5mm}
     \begin{center}
	\includegraphics[width = \linewidth]{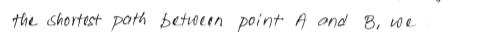}
     \end{center}
	\caption{Sample Single Line Handwritten text Image 2}
	\label{fig: Sample Single Line Handwritten text Image 2}
\end{figure}

\begin{table}
    \centering
        \caption{Results of Single Line Handwritten text Images}
        
    \label{tab: Single Line Handwritten text Images 1}
    \begin{adjustbox}{width=\columnwidth,center}
    \begin{tabular}{|c|c|}
    \hline
     & Image 1 \\ [0.5ex] 
     \hline\hline
      Ground Truth & Machine Learning is an idea to learn from examples  \\
      \hline
       PP-OCR output & Mochint Leorning is on ideo to leorn from exomples \\
      \hline
      Tr-OCR output & Machine learning is on idea to learn from examples \\ 
      \hline
    \end{tabular}
    \end{adjustbox}
\end{table}

\begin{table}
    \centering
    \caption{Results of Single Line Handwritten text Images}
    \label{tab: Single Line Handwritten text Images 2}
    \begin{adjustbox}{width=\columnwidth,center}
    \begin{tabular}{|c|c|}
    \hline
     & Image 2 \\ [0.5ex] 
     \hline\hline
      Ground Truth & the shortest path between point A and B, we \\
      \hline
       PP-OCR output & Hhe Shortest poth betueen point A and B,We \\
      \hline
      Tr-OCR output & the shortest path between point A and B. see... \\ 
      \hline
    \end{tabular}
     \end{adjustbox}
\end{table}

\subsubsection{Dataset 3: License Plate Dataset}
This dataset consists of 209 cropped license plates  (as seen in Fig. \ref{fig: Sample License Plate Images}) using the original bounding boxes and has all the single characters labeled, creating a total of 2026 character bounding boxes. Every image comes with a .xml annotation file.
The outputs of some sample images in Fig. \ref{fig: Sample License Plate Images} are shown in Table \ref{tab: Results on some License Plate Images}.
\\
Using the ultra-weight PP-OCR model pre-trained on English and Chinese languages resulted in a CER of 0.18.
While the Tr-OCR model was fine-tuned on the SROIE dataset having printed text, it resulted in a CER of 0.24.
Hence PP-OCR performed better than Tr-OCR on this dataset.
\\
\subsubsection{Dataset 4: Single Line Handwritten Text Datasett}
This dataset contains handwritten single-line images (as seen in Fig. \ref{fig: Sample Single Line Handwritten text Image 1} and Fig. \ref{fig: Sample Single Line Handwritten text Image 2}), and it's labeled similarly to the IAM dataset. Around 400 lines of handwritten images with their labels are provided.
\\
Using the ultra weight PP-OCR model, pre-trained on English and Chinese languages, resulted in a CER of 0.53 and a WER of 0.8.
While Tr-OCR model pre-trained on the IAM dataset of handwritten text resulted in a CER of 0.09 and WER of 0.24.
\\Hence Tr-OCR performed better than PP-OCR on this dataset.
The outputs of some sample images in Fig. \ref{fig: Sample Single Line Handwritten text Image 1} and Fig. \ref{fig: Sample Single Line Handwritten text Image 2} are shown in Table \ref{tab: Single Line Handwritten text Images 1} and Table \ref{tab: Single Line Handwritten text Images 2} respectively.
\subsection{Classification}
The model to classify the text into handwritten and printed text was tested on 2 datasets i.e. the Bing Images of Short quotes (discussed earlier) and a self made handwritten dataset of around 30 images.
\\In handwritten document dataset, it classified 30 out of 32 images correctly as handwritten text and 2 incorrectly as printed text. In printed quotes dataset, it classified 191 out of 198 images correctly as printed text and 7 incorrectly as handwritten text.
Overall the classification model has an accuracy of about $96\%$.

\subsection{Module-A pipeline Results}
 A mutli-line document is first fed to the segmentation module which breaks the document down to single line texts,  each line is then fed to the classification model which classifies it as handwritten or printed text. If it is handwritten text, the TrOCR model trained on handwritten text is used to perform OCR on it and if it is classified as printed text then the TrOCR model trained on printed text is used to perform OCR on it.
The OCR output for each line is then clubbed and the output corresponding to the input document is obtained.

\begin{figure}
    \begin{center}
	\includegraphics[scale = 0.10]{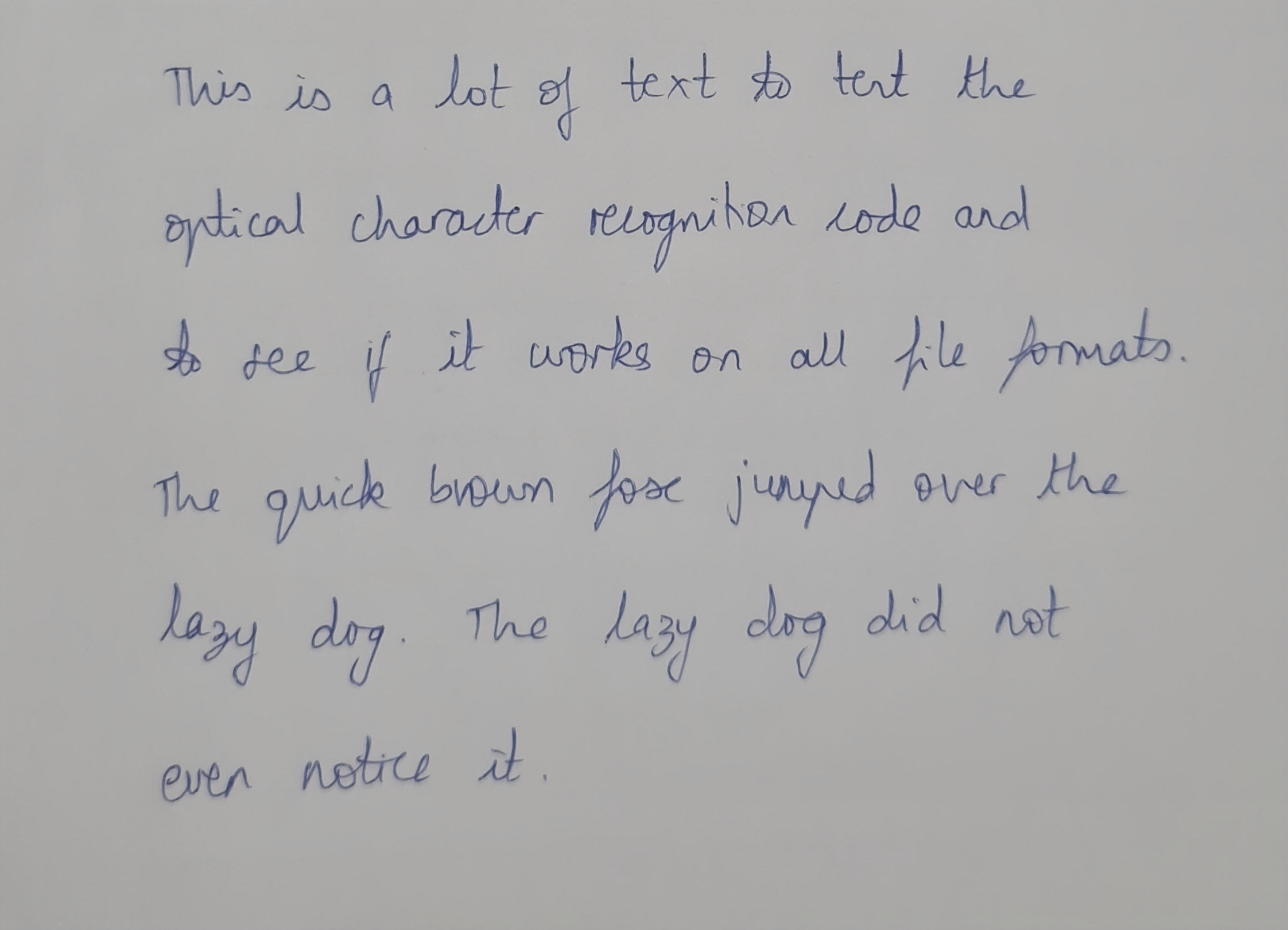}
    \end{center}
	\caption{Input document}
	\label{fig: Input document}
\end{figure}

\begin{figure}
    \begin{center}
	\includegraphics[scale = 0.25]{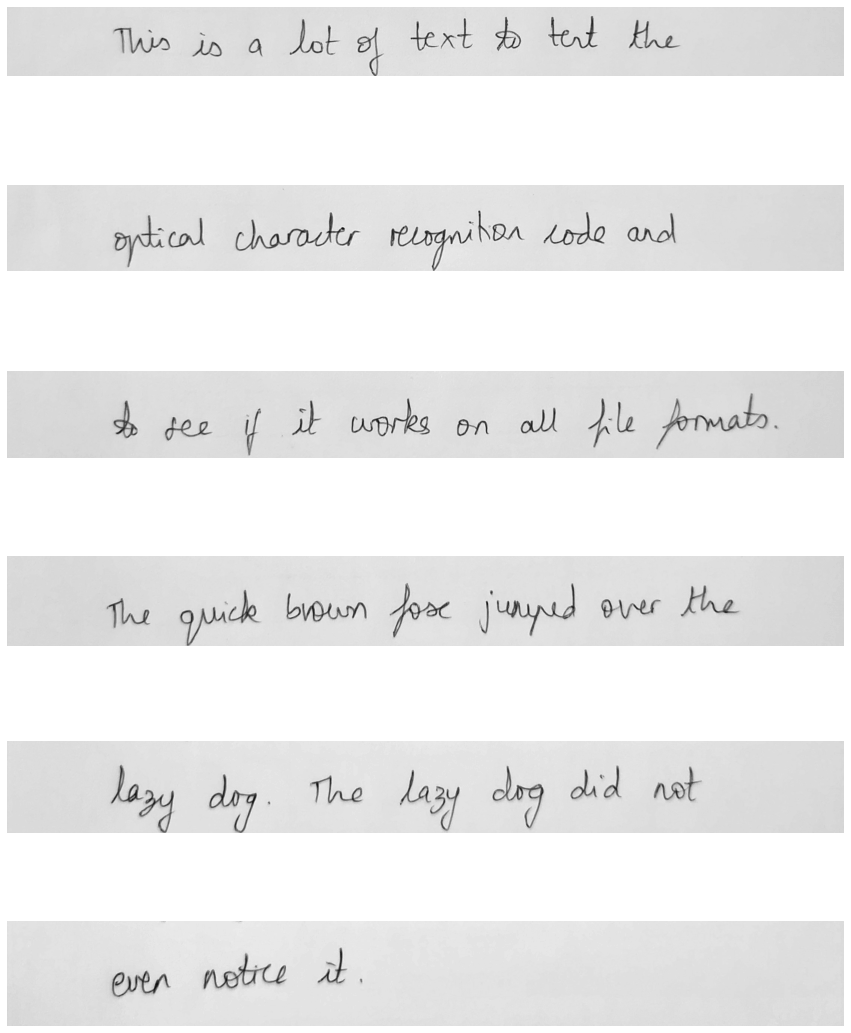}
    \end{center}
	\caption{Input document after segmentation}
	\label{fig: Input document after segmentation}
\end{figure}

Figure \ref{fig: Input document} is an example of a handwritten document.
\\After segmenting it into individual lines we get we get Figure \ref{fig: Input document after segmentation}.
\\The classification model classifies each line correctly into printed text as show in Table \ref{tab: Results of Classification of individual lines of the input document}.
\\We then perform OCR using TrOCR model pre-trained on handwritten text. Results obtained are shown in Table \ref{tab: Results of OCR of individual lines of the input document}.
The CER for this example was 0.079 and WER was 0.2.
Similarly we performed this pipeline over few more examples; both printed and handwritten.
The average CER over all these examples is 0.103 and average WER is 0.274.
\begin{table}[t]
    \centering
    \caption{Results of Classification of individual lines of the Input document}
    \label{tab: Results of Classification of individual lines of the input document}
    \begin{tabular}{|c|c|c|c|}
    \hline
     Line No. & handwritten probabilty & printed probability & Classification \\ [0.5ex] 
     \hline\hline
      Line1 & 0.73 & 0.27 & handwritten\\
      \hline
      Line2 & 0.71 & 0.29 & handwritten\\ 
      \hline
      Line3 & 0.75 & 0.25 & handwritten\\ 
      \hline
      Line4 & 0.77 & 0.23 & handwritten\\
      \hline
      Line5 & 0.74 & 0.26 & handwritten\\
      \hline
      Line6 & 0.55 & 0.45 & handwritten\\
      \hline
    \end{tabular}
\end{table}

\begin{table}[t]
    \centering
    \caption{Results of OCR of individual lines of the input document}
    \label{tab: Results of OCR of individual lines of the input document}
     \begin{adjustbox}{width=\columnwidth,center}
    \begin{tabular}{|c|c|c|c|}
    \hline
     Line No. & TrOCR output & Ground Truth \\ [0.5ex] 
     \hline\hline
      Line1 & This is a lot of text to test the & This is a lot of text to test the\\
      \hline
      Line2 &  political character recognition code and & optical character recognition code and\\ 
      \hline
      Line3 & \# to see if it works on all file formats. & to see if it works on all file formats.\\ 
      \hline
      Line4 & \# The quick brown face jumped over the & The quick brown fox jumped over the\\
      \hline
      Line5 & " Lazy dog. The Lazy dog did not & lazy dog. The lazy dog did not\\
      \hline
      Line6 & seven notice it. & even notice it.\\
      \hline
    \end{tabular}
    \end{adjustbox}
\end{table}

\begin{figure}[!htb]
    \centering
    \includegraphics[width=0.5\linewidth]{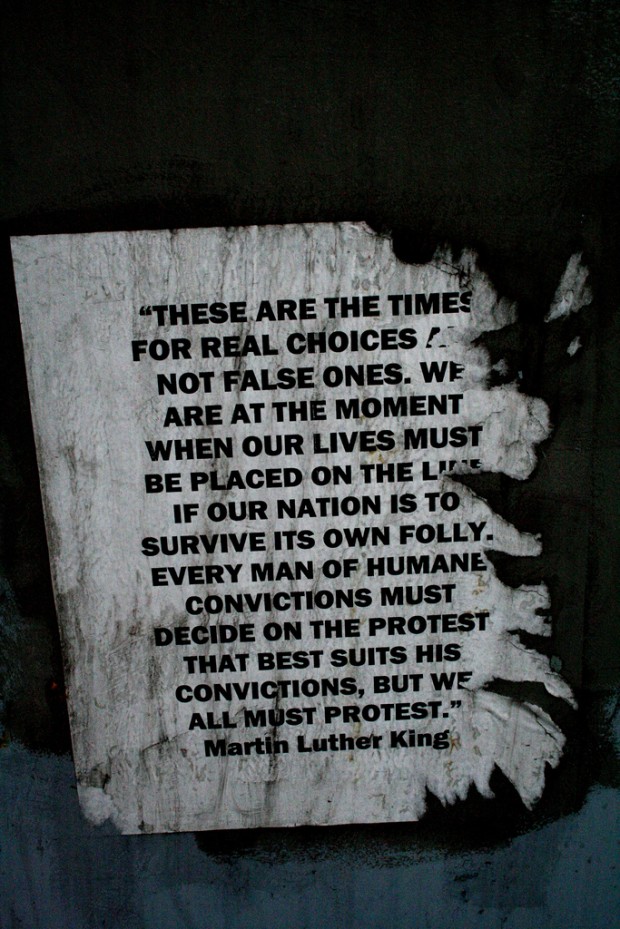}
    \caption{Test 1: A sample one-line quote\\OCR Output: THESE ARE THETIMES FORREALCHOICES NOTFALSEONES.WE AREATTHE MOMENT WHEN OUR LIVES MUST BEPLACED ON THEL IFOUR NATIONISTO SURVIV \\
NLP Output: THESE ARE THE TIMES FOR REAL CHOICES NOT FALSE ONES WE ARE AT THE MOMENT WHEN OUR LIVES MUST BE PLACED ON THE L I FOR NATION IS TO SURVIVE \\
Alpaca-LORA Output:These are the times for real choices not false ones. We are at the moment when our lives must be placed on the four nations to survive}
    \label{fig:example1}
\end{figure}
\begin{figure}[!htb]
    \centering
    \includegraphics[width=0.5\linewidth]{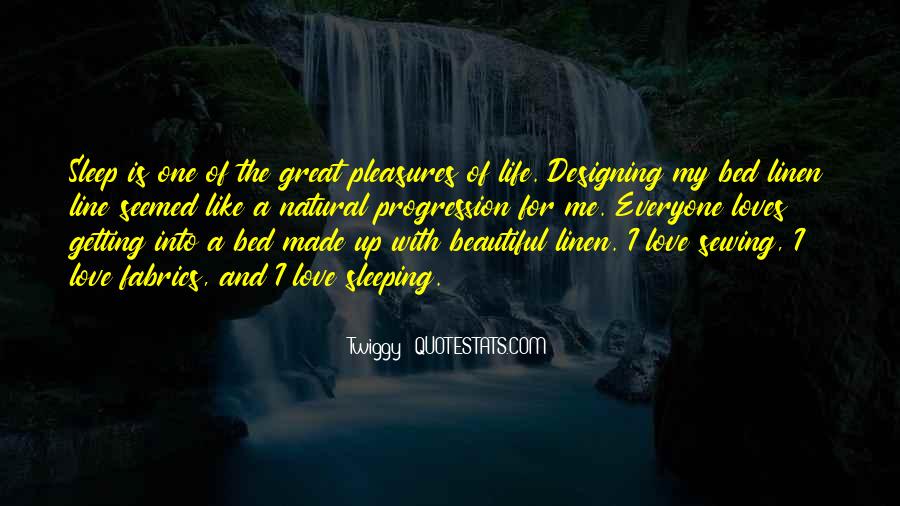}
    \caption{Test 2: A sample one-line quote\\
    OCR Output: Sleep is one of the great pleasures ot life. Designing my
hed linen line Seemed like a natural progression for me. Everyone loves getting into a bed made up with beauitiful linen. 1 love Sewing. love tabrics,and 1 love sleeping 666im QUOTESTATS.COM \\
NLP Output: sleep is one of the great pleasures ot life Designing my
he d linen line Seemed like a natural progression for me Everyone love \\
Alpaca-LORA Output: Sleep is one of the great pleasures of life. Designing my hed linen line seemed like a natural progression for me. Everyone loves getting into a bed made up with beautiful linen. I love sewing. I love textiles, and I love sleeping.}
    \label{fig:example3}
\end{figure}
\subsection{Results after Post Processing}
The figures \ref{fig:example1} and \ref{fig:example3} are two results of our pipeline revealing the Images of One Line Quotes, the OCR Output, and the Post-Processed output.
The OCR Output for Fig.\ref{fig:example1} and Fig.\ref{fig:example3} shows how the spaces and spellings are corrected by the proposed pipeline.
From Table:\ref{table: Post Processing Metrics Synthetic OSCAR Dataset} and Table:\ref{table: Post Processing Metrics Kaggle Handwritten Dataset}, we can see that both the CER and WER on the datasets get reduced to a great extent.
\begin{table}[]
\centering
\caption{Post Processing Metrics Synthetic OSCAR Dataset}
\label{table: Post Processing Metrics Synthetic OSCAR Dataset}
\begin{tabular}{||c | c | c ||} 
 \hline
 Text Corpus & WER & CER \\ [0.5ex] 
 \hline\hline
 Baseline & 0.455 & 0.124 \\
 \rowcolor{chartreuse(traditional)}
 Seq2SeqLM ByT5 & 0.128 & 0.059\\
 \rowcolor{chartreuse(traditional)}
 Seq2SeqLM BART & 0.062 & 0.013\\
 \hline
 \rowcolor{chartreuse(traditional)}
 Alpaca-LORA & 0.045 & 0.005\\
 \hline
\end{tabular}
\end{table}

\begin{table}[h!]
\centering
\caption{Post Processing Metrics Kaggle Handwritten Dataset}
\label{table: Post Processing Metrics Kaggle Handwritten Dataset}
\begin{adjustbox}{width=\linewidth,center}
\begin{tabular}{||c | c | c ||} 
 \hline
 Model & WER & CER \\ [0.5ex] 
 \hline\hline
 
  OCR Output & 0.363 & 0.169 \\
  TextBlob (Using Peter Norvig's Algorithm) & 0.346 & 0.149 \\
  Symmetric Delete spelling correction algorithm(SymSpell) & 0.308 & 0.135 \\
 \rowcolor{chartreuse(traditional)}
  Seq2SeqLM ByT5 & 0.242 & 0.093 \\ 
 \rowcolor{chartreuse(traditional)}
  Seq2SeqLM BART & 0.192 & 0.088 \\ 
 \hline
 \rowcolor{chartreuse(traditional)}
 Alpaca-LORA & 0.135 & 0.023\\
 \hline
\end{tabular}
\end{adjustbox}
\end{table}
\section{Conclusion}
The evaluation of the two OCR models \textit{viz.} PP-OCR and TrOCR over different datasets showed that TrOCR outperforms PP-OCR in all the datasets except the License plate dataset.
A fine-tuning of the TrOCR is required on the License dataset to provide improved results, which can be considered as a future scope. 
Tr-OCR can be used for OCR of printed and handwritten texts as it gives better results in both cases.
The line segmentation module works well for non-skewed documents. For skewed documents, another algorithm has to be developed for segmentation which can be considered as another future scope.
Similarly, our OCR output Post Processing Pipeline effectively reduces the errors in the OCR Degraded text. This observation can be seen in our results, where for the first synthetically generated dataset, the WER of the OCR Output came down from 0.455 to 0.045, and the CER came down from 0.124 to 0.005. Similarly, on the Kaggle Single Line Dataset, the CER decreased from 0.169 to 0.023 and WER from 0.363 to 0.135.
\section*{Acknowledgement}
The authors like to thank the funds received from IITG Startup grant (xEEESUGIITG01349ANRD001) for the research.
\bibliographystyle{IEEEtran}
\bibliography{references} 

\begin{thebibliography}{10}
\providecommand{\url}[1]{#1}
\csname url@samestyle\endcsname
\providecommand{\newblock}{\relax}
\providecommand{\bibinfo}[2]{#2}
\providecommand{\BIBentrySTDinterwordspacing}{\spaceskip=0pt\relax}
\providecommand{\BIBentryALTinterwordstretchfactor}{4}
\providecommand{\BIBentryALTinterwordspacing}{\spaceskip=\fontdimen2\font plus
\BIBentryALTinterwordstretchfactor\fontdimen3\font minus
  \fontdimen4\font\relax}
\providecommand{\BIBforeignlanguage}[2]{{%
\expandafter\ifx\csname l@#1\endcsname\relax
\typeout{** WARNING: IEEEtran.bst: No hyphenation pattern has been}%
\typeout{** loaded for the language `#1'. Using the pattern for}%
\typeout{** the default language instead.}%
\else
\language=\csname l@#1\endcsname
\fi
#2}}
\providecommand{\BIBdecl}{\relax}
\BIBdecl

\bibitem{liang2005camera}
J.~Liang, D.~Doermann, and H.~Li, ``Camera-based analysis of text and
  documents: a survey,'' \emph{International Journal of Document Analysis and
  Recognition (IJDAR)}, vol.~7, pp. 84--104, 2005.

\bibitem{bansal2020building}
S.~Bansal, M.~Gupta, and A.~K. Tyagi, ``Building a character recognition system
  for vehicle applications,'' in \emph{Advances in Decision Sciences, Image
  Processing, Security and Computer Vision: International Conference on
  Emerging Trends in Engineering (ICETE), Vol. 1}.\hskip 1em plus 0.5em minus
  0.4em\relax Springer, 2020, pp. 161--168.

\bibitem{DBLP:journals/corr/abs-2109-10282}
\BIBentryALTinterwordspacing
M.~Li, T.~Lv, L.~Cui, Y.~Lu, D.~A.~F. Flor{\^{e}}ncio, C.~Zhang, Z.~Li, and
  F.~Wei, ``Trocr: Transformer-based optical character recognition with
  pre-trained models,'' \emph{CoRR}, vol. abs/2109.10282, 2021. [Online].
  Available: \url{https://arxiv.org/abs/2109.10282}
\BIBentrySTDinterwordspacing

\bibitem{DBLP:journals/corr/abs-2009-09941}
\BIBentryALTinterwordspacing
Y.~Du, C.~Li, R.~Guo, X.~Yin, W.~Liu, J.~Zhou, Y.~Bai, Z.~Yu, Y.~Yang, Q.~Dang,
  and H.~Wang, ``{PP-OCR:} {A} practical ultra lightweight {OCR} system,''
  \emph{CoRR}, vol. abs/2009.09941, 2020. [Online]. Available:
  \url{https://arxiv.org/abs/2009.09941}
\BIBentrySTDinterwordspacing

\bibitem{norvig2007write}
P.~Norvig, ``How to write a spelling corrector,'' \emph{De: http://norvig.
  com/spell-correct. html}, 2007.

\bibitem{lansley2020seader++}
M.~Lansley, S.~Kapetanakis, and N.~Polatidis, ``Seader++ v2: detecting social
  engineering attacks using natural language processing and machine learning,''
  in \emph{2020 International Conference on Innovations in Intelligent Systems
  and Applications (INISTA)}.\hskip 1em plus 0.5em minus 0.4em\relax IEEE,
  2020, pp. 1--6.

\bibitem{segmentation}
``Segmenting lines in handwritten documents using a* path planning algorithm,''
  \url{https://muthu.co/segmenting-lines-in-handwritten-documents-using-a-path-planning-algorithm/}.

\bibitem{classification}
``Printed vs. handwritten text lines – automatically separated,''
  \url{https://readcoop.eu/printed-vs-handwritten-text-lines-automatically-separated/}.

\bibitem{digital}
``Overview - born-digital images (web and email),''
  \url{https://rrc.cvc.uab.es/?ch=1\&com=downloads}.

\bibitem{ch2017total}
C.~K. Ch'ng and C.~S. Chan, ``Total-text: A comprehensive dataset for scene
  text detection and recognition,'' in \emph{2017 14th IAPR international
  conference on document analysis and recognition (ICDAR)}, vol.~1.\hskip 1em
  plus 0.5em minus 0.4em\relax IEEE, 2017, pp. 935--942.

\bibitem{LP}
``License plate characters - detection ocr,''
  \url{https://www.kaggle.com/datasets/francescopettini/license-plate-characters-detection-ocr}.

\bibitem{single}
``English handwritten line dataset,''
  \url{https://www.kaggle.com/datasets/sushant097/english-handwritten-line-dataset}.

\bibitem{raffel2020exploring}
C.~Raffel, N.~Shazeer, A.~Roberts, K.~Lee, S.~Narang, M.~Matena, Y.~Zhou,
  W.~Li, and P.~J. Liu, ``Exploring the limits of transfer learning with a
  unified text-to-text transformer,'' \emph{The Journal of Machine Learning
  Research}, vol.~21, no.~1, pp. 5485--5551, 2020.

\bibitem{xue2022byt5}
L.~Xue, A.~Barua, N.~Constant, R.~Al-Rfou, S.~Narang, M.~Kale, A.~Roberts, and
  C.~Raffel, ``Byt5: Towards a token-free future with pre-trained byte-to-byte
  models,'' \emph{Transactions of the Association for Computational
  Linguistics}, vol.~10, pp. 291--306, 2022.

\bibitem{lewis2019bart}
M.~Lewis, Y.~Liu, N.~Goyal, M.~Ghazvininejad, A.~Mohamed, O.~Levy, V.~Stoyanov,
  and L.~Zettlemoyer, ``Bart: Denoising sequence-to-sequence pre-training for
  natural language generation, translation, and comprehension,'' \emph{arXiv
  preprint arXiv:1910.13461}, 2019.

\bibitem{ma2019nlpaug}
E.~Ma, ``Nlp augmentation,'' https://github.com/makcedward/nlpaug, 2019.

\bibitem{2022arXiv220106642A}
J.~{Abadji}, P.~{Ortiz Suarez}, L.~{Romary}, and B.~{Sagot}, ``{Towards a
  Cleaner Document-Oriented Multilingual Crawled Corpus},'' \emph{arXiv
  e-prints}, p. arXiv:2201.06642, Jan. 2022.

\bibitem{piantadosi2014zipf}
S.~T. Piantadosi, ``Zipf’s word frequency law in natural language: A critical
  review and future directions,'' \emph{Psychonomic bulletin \& review},
  vol.~21, pp. 1112--1130, 2014.

\bibitem{wordninja}
\BIBentryALTinterwordspacing
 [Online]. Available: \url{https://github.com/keredson/wordninja}
\BIBentrySTDinterwordspacing

\end{thebibliography}
\end{document}